# FORCED EVOLUTION *IN SILICO* BY ARTIFICIAL TRANSPOSONS AND THEIR GENETIC OPERATORS: THE JOHN MUIR ANT PROBLEM

**Alexander V. Spirov[1*], Alexander B. Kazansky[2], Leonid Zamdborg[1], Juan J. Merelo[3] and Vladimir F. Levchenko[2]**

[1]State University of New York at Stony Brook, Computer Science Department and Center of Excellence in Wireless & Information Technology

[2]The Sechenov Institute of Evolutionary Physiology and Biochemistry, 44 Thorez Ave., St. Petersburg, 194223, Russia

[3]Departamento de Arquitectura y Tecnologia de Computadores, University of Granada, Granada, Spain

kazansky@iephb.nw.ru

jmerelo@geneura.ugr.es

lew@lew.spb.org

[*]**Corresponding author:** State University of New York at Stony Brook, Computer Science Department and Center of Excellence in Wireless & Information Technology, Stony Brook University Research & Development Park, 1500 Stony Brook Road, Stony Brook, NY 11794-6040

Tel: (631) 632-5582

Fax: (631) 632-4653

Email: Alexander.Spirov@gmail.com

**ABSTRACT**

Modern evolutionary computation utilizes heuristic optimizations based upon concepts borrowed from the Darwinian theory of natural selection. Their demonstrated efficacy has reawakened an interest in other aspects of contemporary biology as an inspiration for new algorithms. However, amongst the many excellent candidates for study, contemporary models of biological macroevolution attract special attention. We believe that a vital direction in this field must be algorithms that model the activity of genomic parasites, such as transposons, in biological evolution. Many evolutionary biologists posit that it is the co-evolution of populations with their genomic parasites that permits the high efficiency of evolutionary search found in the living world. This publication is our first step in the direction of developing a minimal assortment of algorithms that simulate the role of genomic parasites. Specifically, we started in the domain of genetic algorithms (GA) and selected the Artificial Ant Problem as a test case. This navigation problem is widely known, and possesses a large body of literature. We add new objects to the standard toolkit of GA - artificial transposons and a collection of operators that operate on them. We define these artificial transposons as a fragment of an ant's code that possesses properties that cause it to stand apart from the rest. The minimal set of operators for transposons is a transposon mutation operator, and a transposon reproduction operator that causes a transposon to multiply within the population of hosts. An analysis of the population dynamics of transposons within the course of ant evolution showed that transposons are involved in the processes of propagation and selection of blocks of ant navigation programs. During this time, the speed of evolutionary search increases roughly an order of magnitude. We concluded that artificial transposons, analogous to real transposons, are truly capable of acting as intelligent mutators that adapt in response to an evolutionary problem in the course of co-evolution with their hosts.

`Keywords: evolutionary computations, genetic algorithms, artificial transposons, the artificial ant problem.`

# 1. INTRODUCTION

Many areas of Evolutionary Computation (EC), especially genetic algorithms (GA) and genetic programming, were inspired by ideas from evolutionary biology. However, modern evolutionary biology has since advanced considerably, revealing that Darwinian evolution (microevolution) is apparently a particular case of the greater mechanisms of macroevolution. Many current branches of research in evolutionary computation implement the evolution of mechanisms, such as functional programs, neural networks, decision trees, cellular automata, L-systems, and finite state automata. The recent achievements in genomics seem more appropriate as an inspirational model for these domains than the classical set of Darwinian algorithms [Cf. 2, 24, 26, 29].

This stimulates us to select, formalize and apply to EC new evolutionary mechanisms that may help to simulate the creative, heuristic and self-organizing character of biological evolution ([47 - 56].

We concentrate on the mechanisms of natural genetic engineering, whose key actors are mobile genetic elements [Cf. 40 - 41]. The basic idea can be outlined as follows:

- The genome of every organism has mechanisms for genomic rearrangement;

- These mechanisms are activated during periods of evolutionary crisis;

- The mechanisms cause multiple systemic rearrangements of genomes within a few generations.

The key players of natural genetic engineering are mobile genetic elements (synonymous or related terms are jumping genes, selfish DNA, transposons and retroelements [15, 27, 30, 31]. Many biologists speculate that processes in the world of transposons, existing on a substratum of genomes of the greater biological community, are the main source of ~~macro~~evolutionary creativity [8, 20, 40, 41].

There are several groups of genomic parasites. Transposons form the most sophisticated one. Transposon is a mobile piece of DNA that is flanked by terminal repeat sequences and typically bears genes coding for transposition functions. "…it is now recognized that a significant portion of the genome of any eukaryote is composed of "selfish" or "parasitic" genetic elements, which gain a transmission advantage relative to other components of an individual's genome, but are either neutral or detrimental to the organism's fitness" [15]. Arguments have been made that the structure of eukaryotic genomes, including the abundance of transposons, repetitive DNA and introns, provide high evolvability despite possible detrimental effects on the host [40, 41].

It has been estimated that the majority of the DNA in higher organisms is neither translated into proteins nor involved in gene regulation, and is simply "junk" DNA. The bulk of junk DNA is composed of intermediate-repeats comprised of DNA elements that are able to move (or transpose) throughout the genome. These mobile DNA elements are sometimes termed "selfish", since they do not appear to directly benefit the host. Their behavior is purely parasitic: they jump between different sections of a genome in order to propagate themselves, a behavior that seems at first glance to be usually detrimental to their host. However, they do appear to benefit the host organism by providing genomic rearrangements that permit the evolution of new genetic networks.

Transposons are ubiquitous and may comprise up to 45% of an organism's genome [15, 27, 30, 31]. DNA sequences of transposon origin can be recognized by their palindrome endings, flanked by short, non-reversed, repeated sequences resulting from insertion after staggered cuts. Many transposons have a unique DNA site that acts as a forwarding address, directing the transposon to a complementary site elsewhere in the host genome. There are usually multiple copies of any given DNA site in the host genome, and the transposon will attach to a proper site in a random manner. Transposons may grow by acquiring more sequences; one such mechanism involves the placement of two transposons into close proximity so that they act as a single large transposon incorporating the intervening code.

Data was obtained recently that gives credence to the idea that transposable elements are a major source of genetic change, such as the creation of novel genes, the alteration of gene functions, and the genesis of major genomic rearrangements [11, 15, 27, 30, 31]. The long coexistence of transposable elements in the genome is expected to be accompanied by host - transposon co-evolution. There exists an opinion that transposons should be treated as non-random (or rather "intelligent") mutators. Furthermore, transposons can be considered an evolutionary "SOS-crew", cooperatively acting "in emergency", during periods in which the host is experiencing genomic stress [27, 32].

Modern evolutionary biology and evolutionary genetics (and genomics) amassed a great deal of knowledge regarding the complex mechanisms of genomic rearrangement governed by transposons [1, 8, 15, 20, 30, 31]. From an EC point of view, the most sophisticated of these mechanisms may be treated as a local evolutionary search. By this, we mean the mutational hot spots generated by transposons. A promising approach is to include the world of selfish elements in the standard toolkit of EC, and to use the objects/procedures for maintaining and manipulating transposons as a new branch of EC.

In this article, we apply modern evolutionary ideas to improve the performance of a benchmark test known as the ant problem [17, 21]. Specifically, we enhance standard genetic algorithms by including artificial transposons and genetic operators that mimic some of the essential characteristics of real transposon behavior. Because populational and co-evolutional aspects of transposon activity are thought to be essential for understanding macroevolutionary mechanisms, we study in detail the population dynamics of artificial transposons for our version of the ant problem.

The article is organized as follows. In the next section we briefly describe the state-of-art in the field of biologically inspired EC, in section 1.2 we take a look at the key characteristics of biological transposons as an inspirational model, and section 1.3 contains an introduction to the ant problem. In the first sections (2.1 - 2.2) of "Methods and Approach" we give our definitions of artificial transposons and the operators acting on them, in section 2.3 we introduce a modified version of the GA-ant program worked out by Patrick Brennan, and in section 2.4 we describe the concrete implementations of transposons and their operators developed for the ant program. In section 3.1 of the "Results" we present findings demonstrating the improvement of GA by artificial transposon activity, and in the following sections 3.2 to 3.5 we present findings regarding the populational and dynamic aspects of transposons. Finally, in sections 4.1 to 4.3 of the "Discussion" we discuss the probable mechanisms of improving GA by transposon activity in the case of the ant problem, and in the last section we outline the Future Work.

## 1.1. Biologically Inspired EC

It is felt that the field of EC is attempting to make evolutionary algorithms more effective by applying techniques inspired by the latest achievements in biology. These attempts have been termed *genomic algorithms*, *bacterial algorithms* or even "genetic" (quotes intended) algorithms. Many other techniques taken from biology, such as transposition, host-parasite interaction and gene-regulatory networks have also been applied to evolutionary computation. There is no uniform nomenclature, however, and sometimes the same terms are applied to different methods, or the other way around.

These techniques can be divided into three broad categories:

•*Host-parasite methods*. These methods are based on the co-evolution of two different populations, one of them acting as a "parasite", and the other acting as the "host"; the parasites usually encode a version of the problem domain, and the hosts encode the solution to the problem [14, 18, 34, 35, 37, 38, 39]. These methods have been used mainly to evolve sorting networks, but similar approaches have been used, for instance, to evolve players of the *Othello* game. In this case, the approach is not significantly different from other co-evolutionary approaches, in which solutions and instances of a problem are co-evolved.

•*Transposition operators*. These are sometimes known as "bacterial" algorithms [13, 33, 42 - 46]. The basic idea of this approach is to make intra-chromosome crossovers, that is, the crossover of a chromosome with another part of itself, or else an asymmetric crossover, in which a donor chromosome transfers a part of its genetic material to an acceptor chromosome. In some cases, these operators seem to be better than classical genetic algorithms for combinatorial optimization problems.

•*Other biological approaches*. Luke et al. [29] use a method similar to genetic regulatory networks to evolve finite state automata that represent a language grammar, a class of objects that cannot be easily represented using serial, bitstring, genetic algorithms. Burke et al. [9] try to make the evolutionary process closer to its real molecular genetics base, by having a 4-letter genetic alphabet, transduction processes, and variable-length genetic algorithms. In both cases, it is not a general-purpose technique; they only draw some elements from biology to solve problems that would be difficult to solve in other cases.

So far, no technique has been found in the literature that is general enough to be applied to a wide range of problems, and that, in some cases, is able to yield as good or better results than evolutionary algorithms.

## 1.2. Key Characteristics of Transposons as an Inspirational Model

Transposons act as non-random mutators for the genes of the host. They cause different types of mutations when they transpose themselves. Different mechanisms of transposon activity cause different types of mutations:

1) *Non-random insertions.* The insertion of transposons is in some cases targeted rather than random [30, 31]. Certain regions of the host DNA are favored for insertion; these are referred to as *hot-spots* and tend to be specific to a single transposon or a family of such. The specificity of insertion is based on factors encoded by transposon genes.

2) *Recombinational hot-spots.* Transposable elements serve as recombinational hot-spots, allowing the exchange of genetic material between unrelated chromosome sequences [30]. Because they are so abundant, they can mediate large scale chromosome restructurings by means of homologous recombination between similar transposable elements at distant locations.

3) *Mutational hot-spots.* Transposable elements also serve as mutational hot-spots, allowing an increase in the rate of mutation at the level of single genes. For instance, the two key enzymes that control the generation of an inexhaustible variety of antibodies by the immune system are relics of an ancient transposon [1]. A great deal of antibody variability is due to the way antibody genes are assembled: by joining three separate sequences (denoted V, J, and D), each of which comes in many variants, into a single antibody gene. The enzymes that do this assembly, Rag1 and Rag2, work just like the transposase enzymes that mobilize transposons.

4) *Rearrangements of host genetic networks.* In addition to moving themselves, all types of transposable elements occasionally move or rearrange neighboring DNA sequences of the host genome. It has been estimated that 80% of spontaneous mutations are caused by transposons [30, 31]. Repeated sequences, resulting from the activity of mobile elements, range from dozens to millions of copies per genome. Transposons "strew" the genome with different regulatory sequences. By means of this, transposons can rearrange hosts' genetic networks [30, 31]. Transposons may capture genes and move them wholesale to new parts of the genome. They make possible DNA shuffling that can place genes in new regulatory contexts and, thus, possibly, giving them new roles. But what is more, when two transposable elements recognized by the same transposase integrate into neighboring chromosomal sites, the DNA between them can be subjected to transposition. As far as this process provides a particularly effective pathway for the duplication and movement of exons, these elements can help to create new genetic ensembles.

Transposons co-evolved with their host genome as result of selection favoring transposons, which introduced useful variation through gene rearrangement. "Smart" genetic operators evolved in the same manner [40, 41]. It is likely that transposons represent these highly coevolved genetic operators. The possession of smart genetic operators must have contributed to the explosive diversification of higher organisms by providing them with a capacity for natural genetic engineering.

In designing artificial evolution, it would be worthwhile to introduce such genomic parasites, in order to facilitate the sharing of the code that they bring about. Recently, some authors proposed and tested operators imitating some properties of biological transposons [42 - 46]. The distinguishing feature of our approach is the exploitation of the deeper evolutionary aspects of transposons. An example of this would be the imitation of the co-evolution these genomic parasites experience with the host genomes that they populate. Here and elsewhere we use terms accepted in ecology for the description of host-parasite dynamics and evolution. Every genome, and as such every host, has a set of mobile mutational agents (genomic parasites) of its own. Host and parasite populations co-evolve and sometimes, this co-evolution can take the form of an "arms race" between them.

## 1.3. The Ant Problem

The artificial ant problem is a simulation of ant navigation aimed at passing through the labeled trail (nicknamed "The John Muir Trail" in the UCLA experiment [17]). The problem has been repeatedly used as a benchmark problem [10, 12, 16, 21, 22, 25, 28]. In [17], the artificial ant has to follow the John Muir trail placed on a grid. The ants are simple Finite-State Automata (FSA) or an Artificial Neural Network (ANN) that can move along the grid and test their immediate surroundings. The trail starts off as being quite easy to follow, and gradually becomes more difficult, as the turns become more unpredictable and gaps appear (Fig. 1). Each black (labeled) cell is numbered sequentially, from the $1^{st}$, which is set directly next to the starting cell, through to the $89^{th}$, the last cell. The ant's task is to follow this trail and move across each black cell in sequence.

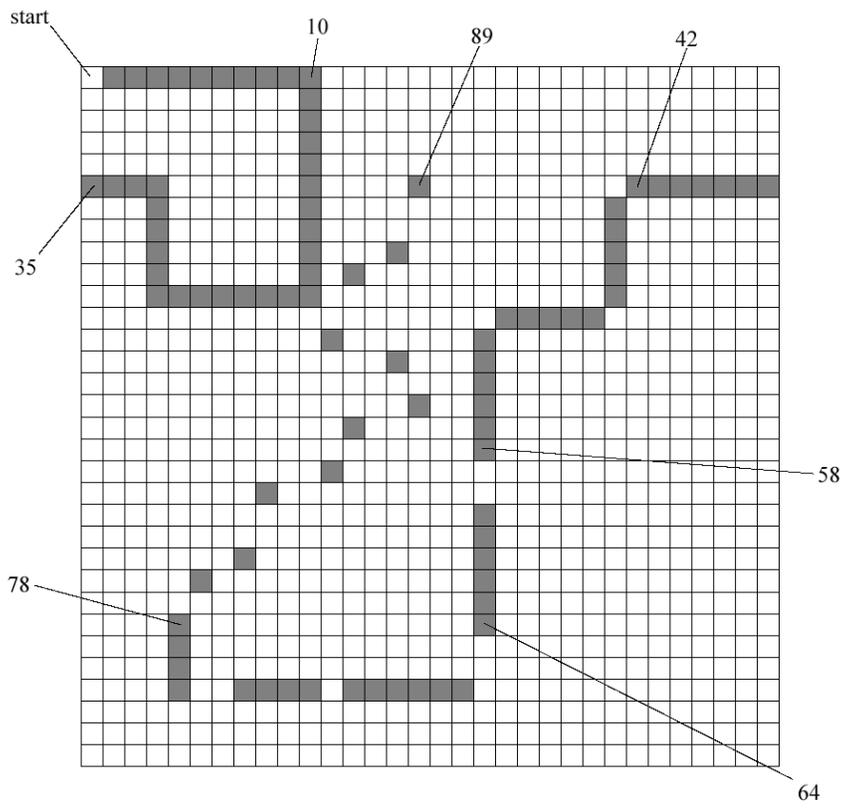

**A.**

**B.**

| State | Input=0 | Input=1 |
|-------|---------|---------|
| 00 | FWD/0A | NOP/09 |
| 01 | RGT/0E | FWD/03 |
| 02 | NOP/08 | NOP/0E |
| 03 | FWD/13 | RGT/18 |
| 04 | NOP/17 | NOP/0B |
| 05 | RGT/17 | RGT/0A |
| 06 | FWD/04 | NOP/09 |
| 07 | LFT/0A | LFT/17 |
| 08 | LFT/12 | FWD/1E |
| 09 | RGT/1E | RGT/16 |
| 0A | RGT/16 | FWD/06 |
| 0B | RGT/13 | NOP/16 |
| 0C | LFT/03 | LFT/0B |
| 0D | LFT/0E | NOP/14 |
| 0E | LFT/0C | NOP/12 |
| 0F | RGT/15 | FWD/1F |
| 10 | NOP/0E | LFT/17 |
| 11 | FWD/12 | FWD/0F |
| 12 | LFT/11 | NOP/0C |
| 13 | RGT/02 | NOP/1D |
| 14 | RGT/0C | LFT/0E |
| 15 | FWD/18 | FWD/09 |
| 16 | FWD/01 | NOP/08 |
| 17 | NOP/0B | LFT/1A |
| 18 | LFT/13 | NOP/11 |
| 19 | NOP/0D | RGT/01 |
| 1A | NOP/1E | LFT/1B |
| 1B | FWD/03 | FWD/10 |
| 1C | RGT/0A | NOP/00 |
| 1D | RGT/06 | LFT/0A |
| 1E | RGT/0C | NOP/18 |
| 1F | RGT/10 | FWD/04 |

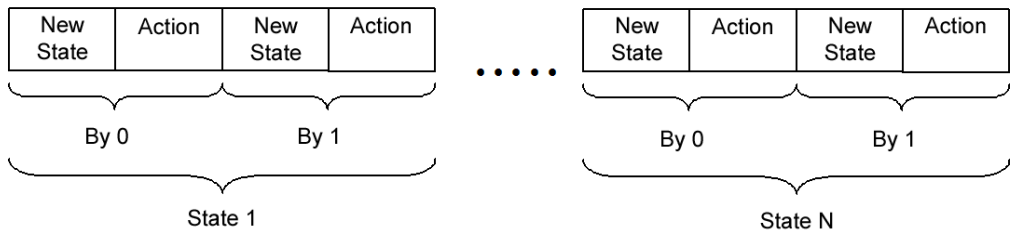

**C.**

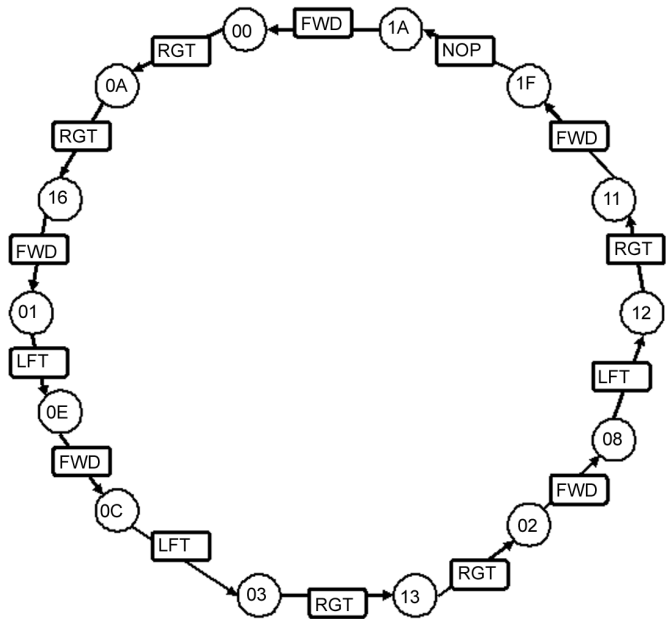

D.

Figure 1. (A). The John Muir Trail. Some "milestone" squares are numbered. The trail itself is a series of black squares on a 32x32 white toroidal (i.e., wraparound) grid. Each black cell is numbered sequentially, from the $1^{st}$ to the $89^{th}$.

(B). A sample implementation of an ant as a finite state automaton. There are 32 states in all. Each state may lead to one of two possible states, depending on the input signal. The input signal is what an ant sees before him. If the cell before him is black then the input is 1, and if white the input is 0. Furthermore, each state transition is accompanied by one of four possible movements (FORWARD - FWD, RIGHT - RGT, LEFT - LFT, or NOP). (See text for further details.)

(C). An integer string with 4*N integers (two links and two actions per state) was used to codify the information contained in the transition table (B).

(D). A graph representation of the dynamics of an example ant program, corresponding to transition table (B) in the case of a constant "1" input. Each vertex consists of a state (in circle) and an action (in rectangle).

The artificial ant problem for two variants of trail (the "Santa Fe" and the "Los Altos") is well tested and discussed in publications. In both cases the trail has been carefully designed in such a way that it would be easy to navigate at the start but as the ant proceeds through the trail, gaps and unexpected turns emerge progressively. By the end of the trail there are more gaps than labeled cells. However, at the beginning, the successful tactic is very simple - only to pick up trail and to move forward. A little further, the ant will need to learn the trick of turning to the right with the trail. Later, it will need to learn the secret of turning occasionally left, and so on.

The ant stands on a single cell and is oriented to the north, south, east, or west. It is capable of sensing the state of the cell directly in front of it. In each time step, the ant must take one of four actions. It may turn left, turn right, move forward one step, or stand still. (When an ant steps onto a black cell, the cell turns white.) After the ant performs its action, it shifts to a new state. Only the internal state of the ant and the state of the cell in front of the ant are used as inputs to the decision table that determines (a) the ant's action and (b) the state it will assume in the next time step. The decision table itself is coded as the ant's genotype (Fig. 1B). The ant's score is the number of cells passed by the ant for the fixed time period.

## *2. METHODS AND APPROACH*

While the ant test was implemented at least in two different C++ libraries [59], we gave preference to the Patrick Brennan version [7]. This "ANT program" was designed in such a way that it isolates, as far as possible, the components of the genetic algorithm from the trail-following experiment and the ant representation. This was a deliberate decision intended to foster an easy transition for the GA code out of the ANT application, and into whatever application that we may find it useful for.

## 2.1. Artificial Transposons

Artificial transposons act as intelligent and sophisticated mutators. They can generate arbitrary procedures of manipulations with the chromosomes of their hosts. In general, these operators can be unitary, binary or plural. Each host has a population of mutators that operate on it. In the simplest case, transposons are the only source of the host's mutations. If a transposon finds a good mutation strategy, then both host and parasite will get a chance for reproduction. In essence, we would have the virtual co-evolution of hosts and their intelligent mutator-parasites.

It is appropriate here to make some notes concerning the terminology. The *Mobile Genetic Element (MGE) technique* comprises procedures for initialization of mobile genetic elements and procedures for operating with these elements. Hereinafter, mobile elements will be referred to as "transposons", whereas the procedures operating on them will be referred to as "MGE operators".

An artificial transposon is a fragment of chromosome which is specially marked, or which has some specific coding characteristics, sufficient for its recognition as such when needed. The code block marked as an artificial transposon still functionally belongs to the host, and is part of the host's genetic code; however this code block is transmittable from host to host and is able to mutate and grow according to certain rules, which differ from the remaining parts of the host's code.

We tested three ways to distinguish transposon code from non-transposon: usage tags (1), marks (2) or definitions (3). To mark the beginning and the end of a transposon's code block, *special tags* in the host's chromosome could be used. While this trick reminds one of the methods used in nature, we found more convenient to use not tags but marks.

In double-string chromosome notation, the additional string will contain blocks of *special marks* that will determine the position of transposon elements. The main string (binary or non-binary) is used for codes, while the additional one (binary) is used for marks. It is possible to use such chromosomes, for instance, in the C++ library Evolving Objects (EO) [19].

Assuming that "1" is the mark of transposon element presence at the corresponding site, and "0" is the mark of its absence, the block (cluster) of marks (...1111111...) distinguishes transposon elements from those of the host. In other words, instead of marking the beginning and the end of a transposon, we mark all its elements in sequence.

The main string containing both host and transposon code blocks can be binary, symbolic, or floating-point. In the case of the ant problem, we can represent our artificial transposon as follows:

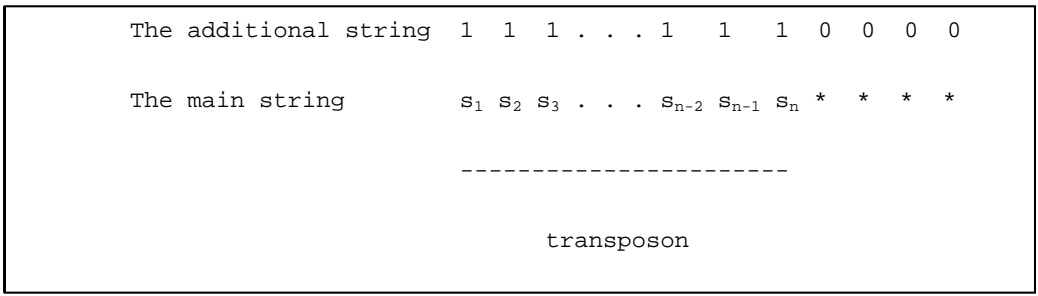

The figure shows the main string, which is symbolic; $s_i$ is a symbol from the alphabet (R, L, F, N); the letters represent ant navigational commands; R means turn right, L - turn left, F means one step forward, and N is NOP.

At the start of computation, an initialisation procedure is used to mark the initial transposons in the initial host population. At first, all elements of the additional string are set to "0". The initialisation procedure makes blocks of marks in this string. The length of a transposon (the number of its elements) can vary from $N_{min}$ ($N_{min} \geq 1$) to $N_{max}$ ($N_{max} \leq CL$, where CL is Chromosome Length). Operators can modify this size through the addition or removal of terminal marks:

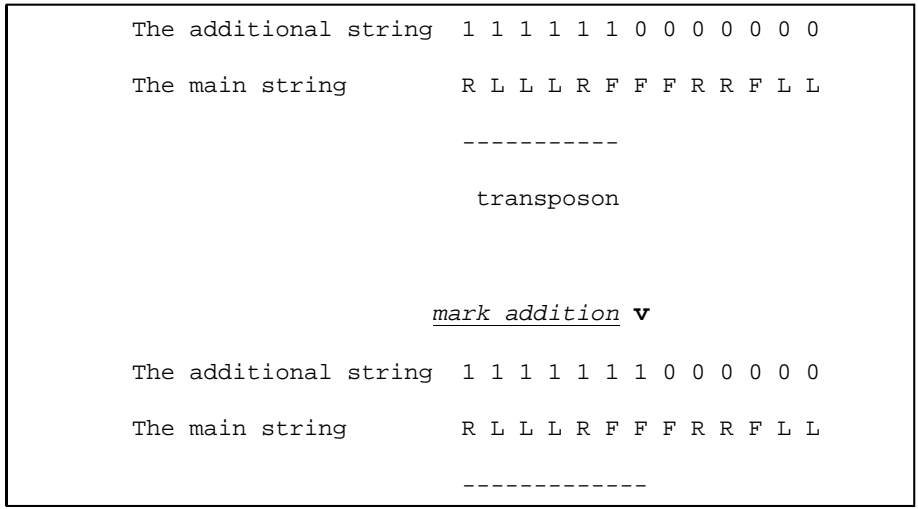

Another possibility to distinguish transposon code from non-transposon is to use a *transposon definition*. The definition of a transposon is a description of its sequence. This definition can be static, i.e. predetermined from the very beginning by the programmer or user, or dynamic, i.e. changing during run-time according some rules predetermined by the programmers. The search procedure called by an operator scans in search of sequences that satisfy this definition; once the sequence is found, the operator performs an action. For instance, one such definition of a transposon sequence is: all palindrome sequences are transposons. In such a case, the search procedure scans chromosomes to find palindromes. If it finds any, then the operator treats that part of the chromosome as a transposon.

Another example of a transposon sequence definition is periodicity of the sequence. That is, let a transposon be any periodic sequence with period N, where $N_{min}$ ($N_{min} \geq 2$) to $N_{max}$ ($N_{max} \leq CL*0.5$, where CL is Chromosome Length). If so, transposon growth will be the growth of the periodic sequence.

A conceptually similar transposon definition was used by Simoes and Costa [42 - 46]. According to that approach, the first (left) flanking sequence of the first transposon in a given chromosome is chosen at random, while the second flanking sequence must be identical or inverse to the first one. The transposon is formed of all the elements between the first element X, leftmost to the first flank and the last element of the second flank.

## 2.2. Operators for Mobile Genetic Elements

The MGE technique includes a set of operators (called MGE operators) for the manipulation of artificial transposons. These genetic operators can be an alternative to, or used in addition to the classical crossover and point mutation mechanism for genetic material recombination. This technique can be applied equally to GA and Genetic Programming (GP), but in this work we concentrated our efforts on testing the applications of MGE operators in GA.

The minimal set of MGE operators should include the two-place transposition operator and the one-place mutation operator. The two-place MGE operator defines the transmission of a transposon from one host to another, thus realizing the reproduction procedure of that transposon in the gene pool of the host population. The one-place MGE operator in the simplest case is a point mutation of the transposon sequence. The characteristics of the specific problem (Royal Roads functions, The Ant problem, TSP, etc.) dictate some modifications of MGE operators.

## 2.3. ANT Program

Patrick Brennan's ANT program [7] is a PC version of the Join Muir Trail experiments performed by the UCLA team [17]. In the original UCLA experiment, a very large population of ants was used, and the ants were alternatively typed as FSAs or ANNs. For each type of ant, a genome encoded the relevant parameters of the ant, either in the form of a state transition table for the FSAs, or connection weights for the ANNs.

Brennan's ANT program duplicates the part of the John Muir Trail experiment that uses FSAs [7]. The program consists of three major subroutines: *Expose*, *Select*, and *Reproduce* (Fig. 2).

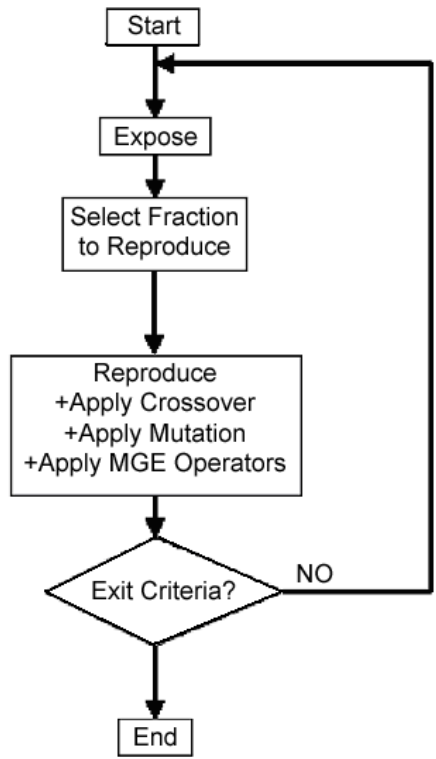

Figure 2. Flow diagram for the ANT program.

In the *Expose* subroutine, each ant in the population is run against the John Muir Trail. Each ant's score is recorded.

In the *Select* subroutine, statistics are generated for the previous Expose run. Each ant's score is compared to the maximum score attained in the population. One of two selection strategies is employed to choose a given ant for reproduction. If the user has selected a *truncation* strategy, only the ants with the highest scores are marked for reproduction. If the user has selected a *roulette-wheel* strategy, an ant with a higher score has a greater chance of being marked, but there is a chance that even the highest-scoring ants may not be marked.

In the *Reproduce* subroutine, the genes of those ants that are not marked for reproduction are overwritten by copies of the genes of those ants that are, and then crossover and mutation are applied.

## 2.4. MGE technique for the Ant Problem

We developed certain new operators, performing processes of replication, mutation and invasion of transposons into specific sites on chromosomes, as well as interactions of transposons with the chromosome (interrelations of the parasite-host type).

## 2.4.1. Artificial Transposons

The FSA implementation of the ant problem is characterized by a complicated genotype-phenotype mapping. The ant genotype is a binary, integer, or symbolic string, while the ant's phenotype is based on the decision tree encoded by the chromosome. In turn, the table can be represented as a state graph (with nodes being states and edges being transitions). Whereas the chromosome is linear, the corresponding transition graph in the general case is non-planar.

Similar implementations are used for such evolutionary tests as The *Tartarus*, The *Bulldozer* and The *Busy Beaver* [3, 36, 57, 58].

Let us recall (See Fig. 1B-C) that the ant binary string-chromosome is coding the state transition table of a FSA. Altogether the automaton has 32 states, ranging in numerical order from state #0 up to state #31 (Fig. 1B). As a number of other authors have used this number of states, we've used it as well, in order to permit a comparison of results [7, 17]. All operators start reading and interpreting the table from state #0. For example, state #0 determines one of the four actions or instructions, and the number of the next state, depending on the binary input value ("0" or "1"). This finite state automation can be represented as a state transition diagram and interpreted as a decision tree but, as far as references to already passed-by states are permissible, the tree can have loops.

At the same time, transposons, both in nature and in our approach, are associated with string-chromosomes. We could use our artificial transposons to manipulate the ant's chromosome, but it would not be a good analogy to transposon-chromosome interaction in nature. In reality (and we accentuated this in the Introductory sections), the interaction of transposons with their host's chromosomes includes the search and recognition of specific sequences. In nature transposons scan chromosome sequences in a search of target signs and a transposon's action is aimed at changes in host gene functions favorable for the transposon. That is why we found it more promising to use transposons to manipulate the state-transition graph (Fig. 1D). By definition, while it is impossible to convert such a graph to a linear sequence of states or actions, it is possible to formulate a set of rules to extract such a sequence from the graph. We can imagine many such sets of rules.

One of the sets we found suitable for our purposes consists of the following rules. Let the first element of the extracted string be the ant's action corresponding to the transition from state 0 under condition input=0. The next state will be state *I*. The second element of the extracted sequence will be the action corresponding to the transition from state *I* under condition input=0. These procedures repeat until the sequence traced forms a loop. The extracted sequence can have different possible lengths but will have two possible forms (for simplicity we use alphabet letters, from A to Z):

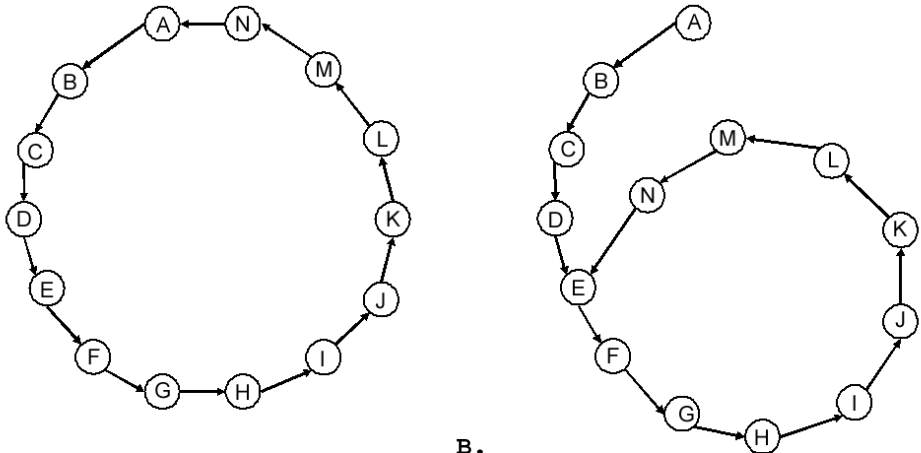

**A.**                              **B.**

These symbol strings serve as "chromosomes" for our transposons, which they manipulate. Each time a MGE operator is applied to a given ant's chromosome, an extraction of the sequence of actions take place. As a result of the MGE-algorithm's operations, the action string is changed. That is why after these changes, another procedure performs the inverse task to write the changes back into the chromosome.

In other words, we take into consideration the half of the decision tree that corresponds to a zero binary input value, i.e. when the ant sees a white (empty) square in front of it. Namely, this part of the ant's transition state governs its search for black squares. In so doing, we reduce the ant's genome to a symbolic string and substantially simplify the following treatment of the problem. An example of such a (closed) symbolic string is:

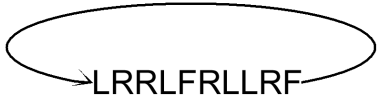

where L is LFT, R is RGT, and F is FWD. The navigation program encoded by this string is optimal for classic Santa Fe trail (Fig. 1A).

It is remarkable that these strings are really variable-length, because they are cycles, but the maximum length is 32 symbolic elements. For instance, the string is shorter but functionally equivalent to the previous one, because it governs scanning of the same three neighbor cells and moving one step forward.

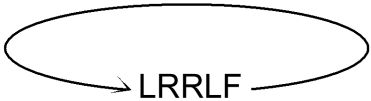

The aforementioned examples are good as navigation programs for our trail as well, but only up to the $64^{th}$ element, and are as such examples of a beneficial transposon (See Fig. 3A). Hence we can interpret these programs as building blocks for search of more sophisticated ant's navigation algorithms.

For this work we use a definition of a transposon (See section 2.1 for details). Namely, for our version of the ant problem, a transposon is a sequence of symbols having the following properties (the definition of the transposon sequence):

1. The sequence should include M symbols (M≤CL, where CL is Chromosome Length).
2. The sequence (excluding the last element) should not contain NOP elements and internal cycles.
3. The sequence should finish with a NOP command or with a reference to one of the states in this sequence, i.e. an internal cycle.

Examples of symbolic strings with transposons that match this definition are:

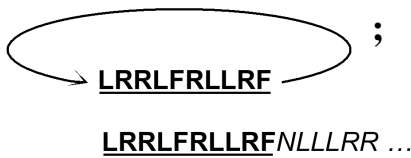

**LRRLFRLLRF**<i>NLLLRR …</i>

The parts of these two symbolic strings proper to our transposon definition are underlined.

During each iteration of the ANT program's execution loop, apart from the MGE operators, the standard mutation and crossover operators act on an ant's binary chromosome as it was done in the Brennan source code. We don't pay any special attention to the "black" half of decision table (input = 1; See Fig. 1B): this part of the decision table undergoes the action of the standard crossover and mutation operators only.

## 2.4.1.1. Transposon-caused mutagenesis

Transposon activity is interpreted as mutational/recombinational hot spots, in accordance with all that is known about host genome rearrangement by such (See the section 1.2). Regardless, one of the very special features of transposons is that they are jumping hot spots. Because of this, artificial transposon activity is a keen example of local searching within a subset of the global GA search. Namely, the initial population of artificial transposons is subjected to mutation and disseminates short code blocks, until they find the first building block. The transposons that cover this building block (for example as it is the case of above-mentioned "R L L F" loop) would have a selective advantage, spread, and become dominant in the host population. In the long run, another transposon will find one more, or larger, or better building block and will receive selective advantage, etc.

Obviously, the simplest way to perform a local search such as this would be the general scheme of random point mutagenesis of transposon sequences coupled with horizontal transmission of transposons, as mentioned in section 2.2.

If we use a definition to find transposon sequences, we have to also define rules for changing the size and code of a transposon. It would be easier if these rules change a transposon's code in such a way that its code still satisfies the definition. For example, we can substitute any symbol in a transposon's code to another one from the list L, R, F:

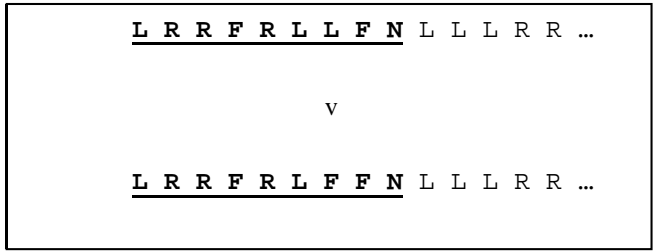

This would be a type of point mutation for transposon code.

Our preliminary test shows that such an algorithm of transposon mutagenesis makes some improvements on ant learning abilities (results not shown). However, in order to study in detail the population dynamics of artificial transposons as sophisticated mutators, we decided to maximally restrict the variability of transposons.

Namely, transposon mutagenesis was targeted not on all transposon elements, but rather at the rightmost element of any given transposon. Hence, only the last element of the sequence undergoes changing. However, apart from the standard point mutation algorithm, the result of this changing is not truly random. In reality, it is an example of constrained or context-dependent point mutagenesis.

For the purposes of context-dependent mutagenesis, we use the idea of the periodicity of transposon sequences (See Section 2.1). Specifically, the action of the last transposon element is substituted for the action of the $N^{th}$ element of the sequence, counted in reverse order. Tested N values ranged from 3 to 8. To our surprise, this very restricted and targeted form of mutagenesis appears sufficient to improve the ant's evolutionary learning. In the rest of the article we will use $N = 5$. The number 5 is not "magic" and our preliminary tests demonstrated that $N = 4$ and $N = 6$ give compatible effects (data not shown). Using this type of transposon definition drastically restricts a transposon's search space and facilitates monitoring of transposon population dynamics. By applying this restrictive definition we exploit a very small part of the manifold of all possible transposon sequences. We tested this restriction and found it to be appropriate for our purposes. As a secondary benefit, this definition makes transposon marking unnecessary.

### 2.4.1.2. Transposon growth

This implementation of our approach does not contain any special algorithm for transposon growth. Our preliminary tests show that the accepted definition of a transposon and targeted mutagenesis on its rightmost element is sufficient to provide growth of transposon sequence length.

If we substitute the terminating NOP element with an L, R or F, then in most cases the code is changed in such a way that its transposon properties are lost. However, this procedure also gives a chance for the transposon to grow:

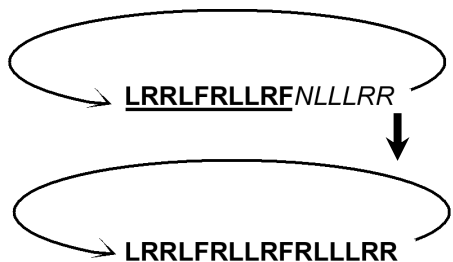

The parts of these two symbolic strings proper to our transposon definition are underlined. As shown, this technique provides an opportunity for the transposon to absorb more elements and grow larger.

Because our definition allows two cases of transposon sequences we found it convenient to distinguish them by name. If the fitting sequence is a cycle, it will be referred to as a *mature transposon*. However, if the fitting sequence terminates with a NOP, we will call it an *immature transposon*. Examples of mature and immature transposons are given on Fig. 2.

**A.**

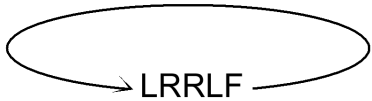

**B.**

| State | Input=0 |
|---|---|
| **0** | **LFT/#17** |
| **17** | **RGT/#13** |
| **13** | **RGT/#21** |
| **21** | **LFT/#9** |
| **9** | **FWD/#0** |

**C.**

**L F F L N** * * * *

**D.**

| State | Input=0 |
|---|---|
| **0** | **LFT/#17** |
| **17** | **FWD/#13** |
| **13** | **FWD/#21** |
| **21** | **LFT/#9** |
| **9** | **NOP/#30** |

Figure 3. Examples of artificial transposons.

(A). A mature transposon. The transposon is a closed five-element cycle of state transitions (0, 17, 13, 21, 9, and again, 0).

(B). The state table corresponding to transposon (A).

(C). An immature transposon. It differs from the previous one by the presence of a NOP action in the last element and is not a cycle.

(D). The state table corresponding to transposon (C).

## 2.4.2. MGE operators

MGE operators scan the predetermined quota of chromosomes in a population. Successively decoding binary chromosome records, this operator seeks sequences that are identified as transposons (mature or immature).

The two-place MGE operator allows the transmission of a transposon from one ant to another, thus implementing the reproduction procedure of that transposon in the gene pool of the host (ant) population. This procedure performs the following operations.

First, a pair of ants is chosen at random. Then, one of them has their chromosome scanned in search of a transposon. If a transposon is found, it is replicated in the other ant's chromosome in the same position as in the source, irrespective of the current content of that location in the chromosome. The chromosome scanning starts from line zero (state#0) and goes on until the first transposon is discovered, at which point scanning ceases irrespectively of the content of the remaining, un-scanned chromosome portion. If no transposons are located, scanning terminates only when the chromosome record ends.

The one-place MGE operator is a type of point mutation, realized under particular conditions. We define this as an *intelligent mutator*.

In detail, the operator acts in the following way: if it finds a sequence in the predetermined length range, terminated by a NOP, then this instruction is substituted for one of the three other actions (FWD, RGT or LFT). Specifically, this NOP is substituted for the action from the $N^{th}$ element of the sequence ($N \in 3,4,5,6,7,8$), counted in order. In all numeric tests described in this article we used N = 5. Presented is a fragment of the output file generated by our ant program in verbose mode that illustrates this kind of transposon mutation:

```
-Check Ant #79 to pattern. Find pattern with NOP. Pattern is:

state #0...LFT/#4

state #4...RGT/#20

state #20...LFT/#23

state #23...LFT/#21

state #21...NOP/#18

Change NOP to node #0 action

state #21...LFT/#18
```

However, if the located sequence is terminated by a reference to the one of the elements inside the sequence (internal cycle), then the MGE operator performs following operation: the action of this element is substituted for the action of the $N^{th}$ element (N = 5), counted backward from the end, the reference being substituted for found at random reference to the element outside of the sequence. One more fragment from output file illustrates this kind of transposon mutation:

```
-Check Ant #51 to pattern. Find pattern with cycle. Pattern is:

state #0...FWD/#25

state #25...FWD/#1

state #1...FWD/#10

state #10...FWD/#21

state #21...FWD/#7

state #7...FWD/#0

Change last action to node #25 action: state #7...FWD/#0
```

The following remark should be noted here. The reference from the last symbol to the first one (i.e., by definition, a mature transposon proper) is processed by the mutator using the same rules. In other words, the mutator breaks the cycle by substituting the reference from the last state to the first one with a reference from the last one to any other, thus generally destroying the transposon.

However this mutation allows the possibility for the transposon to grow as described in Section 2.4.1.2. As such, the action of an intelligent mutator results in the accumulation of large length transposons in the population.

We should emphasize, however, that the two-place MGE operator operates only with mature transposons (See Fig. 3).

## 3. RESULTS

The test trail used in this work is illustrated in Fig. 4. It can be seen that up to the $64^{th}$ element our trail coincides with the Santa Fe trail, but afterwards includes chaotically scattered elements of high difficulty. Being trained on the much simpler early trail portion, the ant is not prepared to surmount the subsequent, complicated sector. More specifically, problems arise at attempts to get over gaps between the $64^{th}$ and $65^{th}$, or the $67^{th}$ and $68^{th}$ cells.

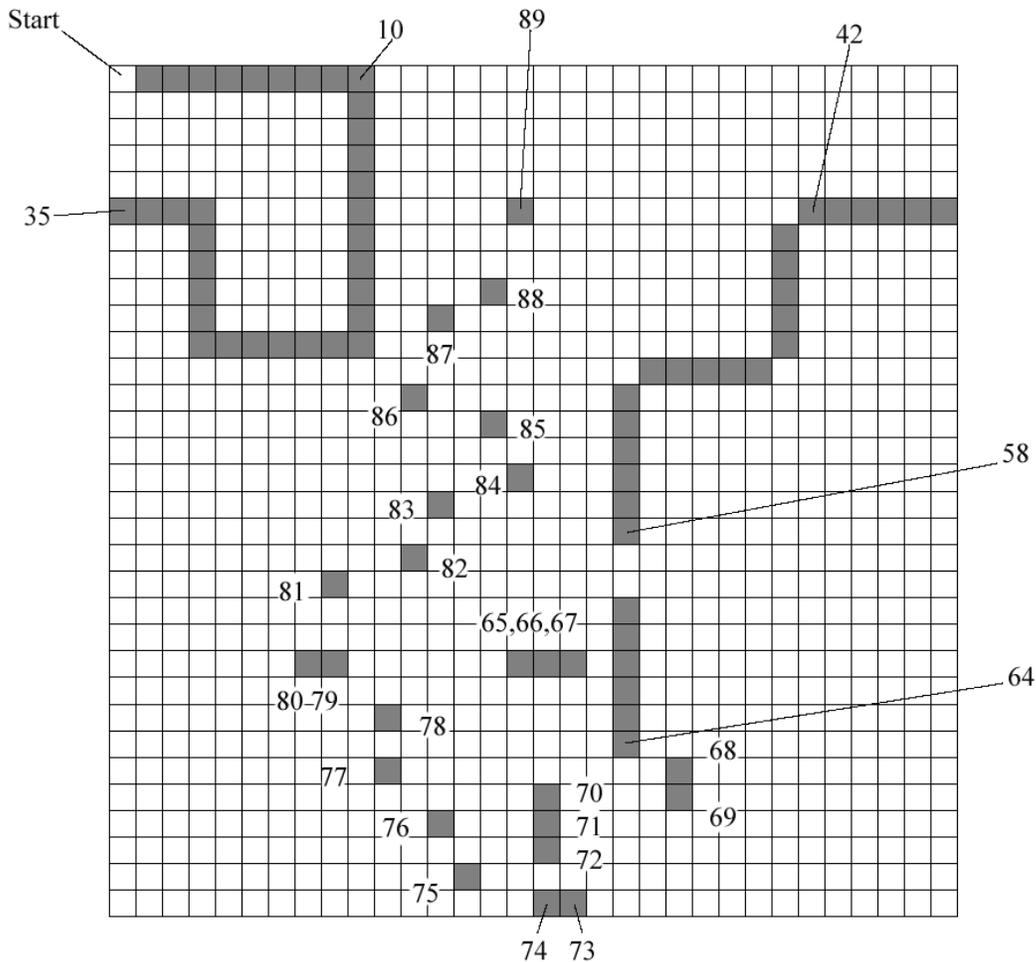

Figure 4. Ant trail used in our experiments. The trail itself is a series of squares on a 32x32 white toroidal grid. Each cell is numbered sequentially, from the $1^{st}$ to the $89^{th}$. The first higher-complexity gap is between the $64^{th}$ and $65^{th}$ cells, and the second is between the $67^{th}$ and $68^{th}$.

## 3.1. Transposons Accelerate the Evolutionary Search

The accelerating effect of transposons is especially noticeable in small populations, where the probability of finding an effective navigation algorithm by applying standard crossover and mutation operators is low.

On this basis, the following experiments were carried out on populations of 100 ants. The choice of such a small population can also be explained by our aim, to carry out a comprehensive analysis of transposon dynamics. Such an analysis is not feasible for large populations of ants because of the great number of transposons. For example, thousands of transposons appear in a population of only 100 ants for 1000 generations.

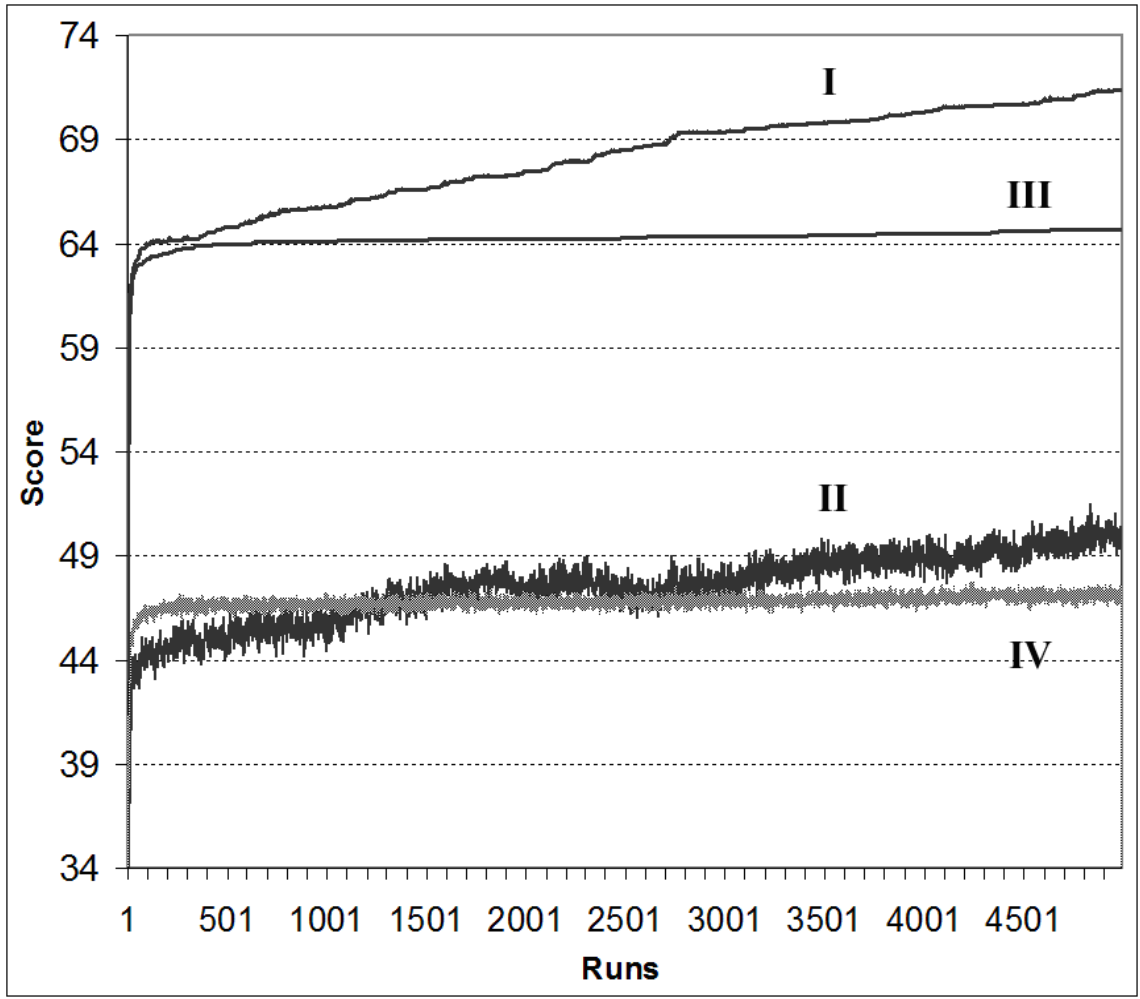

Figure 5. Numerical experiments, statistically demonstrating the increase of GA efficiency due to the effect of MGE operators. A comparison of the mean and best-of-generation score dynamics in the experiment (MGE operators active) with those in the control (MGE operators disabled). The score values are averaged over 100 runs in both cases.

The size of population = 100; the number of generations = 5000; the transposon sequence size varies from 5 to 11; crossover rate (P/bit)/generation = 0.0001; mutation rate (P/bit)/generation = 0.04; **I** are the best-of-generation scores and **II** are the mean scores for the test runs; **III** are the best-of-generation scores and **IV** are the mean scores for the control runs.

Besides limiting the value of ant population abundance, the following parameters were fixed in all test and control runs: the maximum number of an ant's steps (330) and the percentage of the population permitted to reproduce (15%). The *truncation strategy* of reproduction was used when copies of chromosomes with scores exceeding the average value replaced all chromosomes having a score less than the average. An analysis of preliminary runs showed that the experimental results concerned and the inferences made are not sensitive to the variation of these three parameters, i.e. the parameters were not critical in this case.

With the aim of demonstrating the efficiency of the MGE-technique, we performed 100 independent runs of the program, for 5000 generations each. The results of the test and control runs (populations with transposons and without transposons respectively) were compared in several series with the different values of standard mutation parameters. Everywhere in this section we will accept that an effective navigation algorithm should exceed a score of 64 in 330 time steps. Different populations reached different maximum scores in 5000 generations. But the best result was a score of 81, reached in 330 time steps (Cf. Fig. 4).

The results of program runs with the MGE operator and without it are illustrated in Fig. 5. It can be seen that the MGE-technique obviously increases the probability of finding an effective navigation algorithm for small populations and for a low number of generations.

As it is evident from the graphs on Fig. 5, the mean and the best-of-generation score in both experiment and in control were growing roughly linear in time. But the increment of growth in the experiment with transposons was substantially higher than in the control.

It may be suggested that MGE operators raise ant variability primarily in a nonspecific manner, thus supplementing the mutation effect of standard operators. However, this suggestion is not substantiated by the detailed analysis of the mutation process (see below). We carried out two experimental runs with the frequency values of standard mutations being set to two and ten times the value used in the control series given in Fig. 5, respectively. Hence, the high level of standard mutation does not raise the efficiency of the navigation algorithm search; moreover, it decreases this efficiency (the results are not shown graphically).

If MGE operators were disabled, there was no pronounced effect on evolutionary search efficiency from any mutation value or crossover parameter combination. Specifically, the following combinations of crossover and mutation parameters were tested: 0.0001 and 0.01, 0.0 and 0.04, 0.0001 and 0.08, 0.001 and 0.04, respectively (the results are not shown graphically).

## 3.2. Analysis of Transposon Population Dynamics

The transposon population dynamics were studied for a host population of 100 ants, with the duration of the experiment being limited to 500 generations. These limitations were non-critical for the experiment. At the same time, they made possible the graphical representation of the results, because the number of new abundant transposon forms in every run did not exceed 200.

Every generation of a population of 100 ants produces new transposon offspring, consisting of tens of new mutant forms. We accounted for only forms with abundance, exceeding the minimal threshold of 10 individuals. So, for the first 500 generations we had on the order of 100 - 150 new forms.

We performed a comparative analysis of the dynamics of the 25 ant populations which had succeeded in finding an effective navigation algorithm in the predetermined number of generations with the 25 unlucky ant populations that could not discover such an algorithm. Some regularities in the dynamics of ant and transposon populations were revealed. These regularities are illustrated in Fig. 6 and summarized as follows.

1) As a rule, only one form of transposon dominates in a population at any given moment; tens of new forms appear in every host's generation and disappear in at maximum, several generations.

2) From time to time, the current dominating form is displaced by another one. In a large number of generations the former dominating form can reestablish itself (at least, some cases of this were observed).

3) As a rule, 2-3 (less often, 5-6, sometimes up to 10) dominant forms have an opportunity to replace each other in a population for the chosen period of time (the first 500 generations).

4) From time to time, an explosion in the abundance of subdominant forms does occur, which can last for 5-10 generations. Eventually, these subdominant forms can give 2-3 population explosions. One of such explosions can end with the conversion of a subdominant form into a dominating one.

These regularities are inherent equally in the dynamics of stagnant ant populations and in high-evolvable populations, which found an effective navigation algorithm in the period being studied. It turned out that the dynamics of stagnant transposon populations demonstrate the same four properties as those of the ant populations that succeeded in finding an effective navigation algorithm. The moment of finding that algorithm corresponded with changes in the dominant forms.

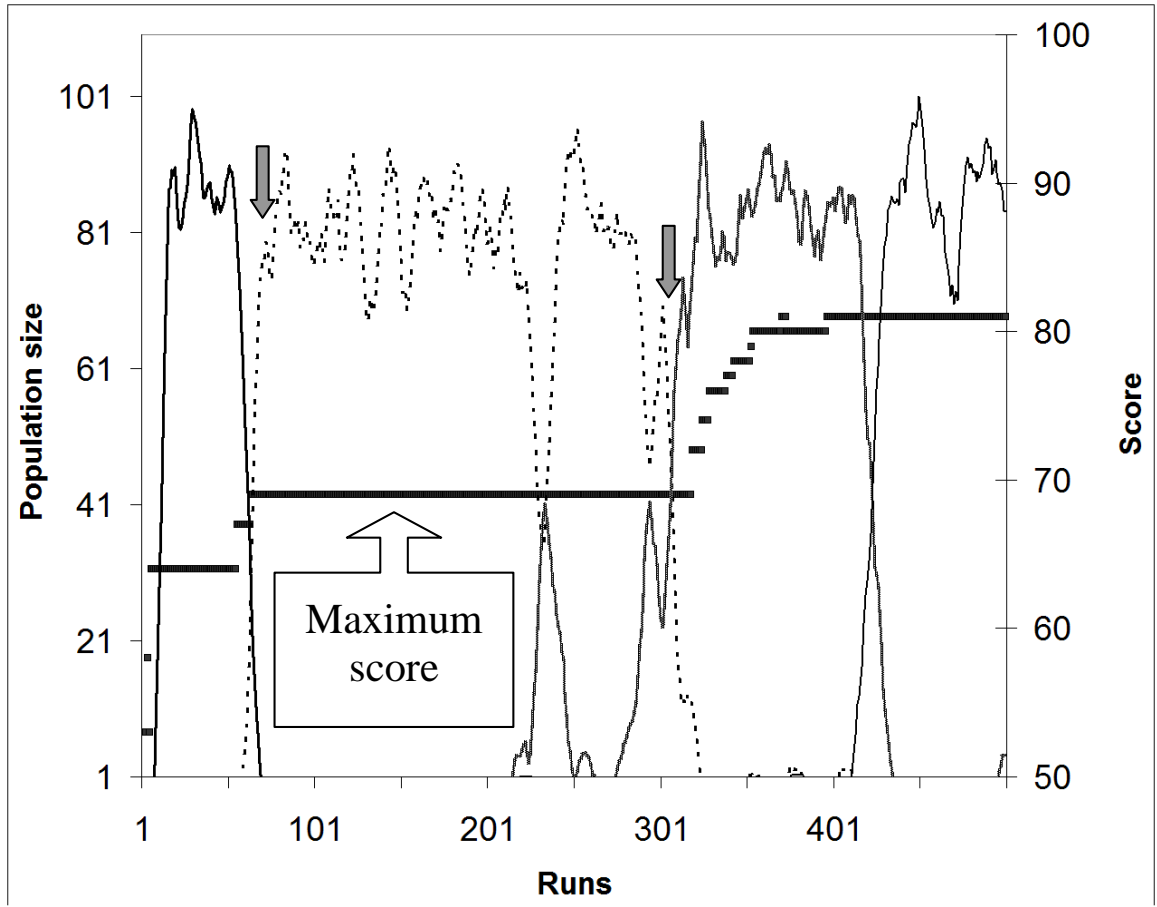

Figure 6. An example of the population dynamics of transposons inside the gene pool of the host (ant) population for the first 500 generations. The curve of the maximum score dynamics for the ant population is given at the top of the diagram (marked with a sequence of squares; the additional score scale is given on the right of the diagram). Moments in which significantly more effective navigation algorithms emerged, and the corresponding shifts in the population's dominant transposon form are marked with downward-pointing arrows (see text for details). The solid, dashed and dotted lines correspond to different dominant transposons.

A comparison of the dynamics of learning in ant population with the dynamics of transposon populations revealed apparent "coincidences" of the moment of change in the dominant transposon form with the moment of finding the first effective navigation algorithm (Fig. 6). Just after this moment, the mean population score steeply rises up to the next plateau. The process lasts for several generations. This local S-shaped function of the mean score growth corresponds with the curves of the population size dynamics of dominant transposon forms (first, the old form, then the new one) (Fig. 6). It should be emphasized, however, that this is only a correlation, *not* a functional relationship. However, the existence of a direct functional relationship between the code of a new dominant transposon form and the code of an effective navigation algorithm remains to be proven.

## 3.3. Larger Transposons Have a More Pronounced Effect on the Rate of Ant Population Training

It was mentioned in the section "Methods and Approach", that by default, the MGE operator is working only with sequences of lengths that lie in the range from 5 to 11. In all experiments, illustrated in the Figs. 5-6, the admissible sizes of transposons lie in the range from 5 to 11. Analyzing transposon population dynamics, we revealed that very rapidly, over some tens of generations, all transposons reached the upper length threshold (= 11). Moreover, sequences of length 12 and more appear in the population. The MGE operator recognizes these large sequences but leaves them intact, i.e. it does not apply the mutation procedure to them. This is explained by the fact that the one-place MGE operator causes mutation not only in the sequences close to transposons by the definition (See "MGE operators" section), but in mature transposons as well. Selection pressure causes surprisingly rapid elimination of all transposons of length less than 11 (irrespectively of maturity or immaturity) and an accumulation of sequences of length 12 and more (the MGE operator does not act on the latter).

That is why we also investigated the behavior of an ant population having transposons of larger lengths. It became clear that the revealed regularity viz., the domination of transposons of maximum length, is valid for all transposon sizes up to 32 (the physical limit). Moreover, it turned out that transposons with a higher upper size threshold have a more pronounced effect on the rate of ant training. This was determined when the averaged score dynamics, being taken from the previous experiments (see Fig. 5) were compared with the score dynamics obtained in an experiment with the same parameters as the previous except one - the upper transposon length threshold which was made equal to 32 (Fig. 7).

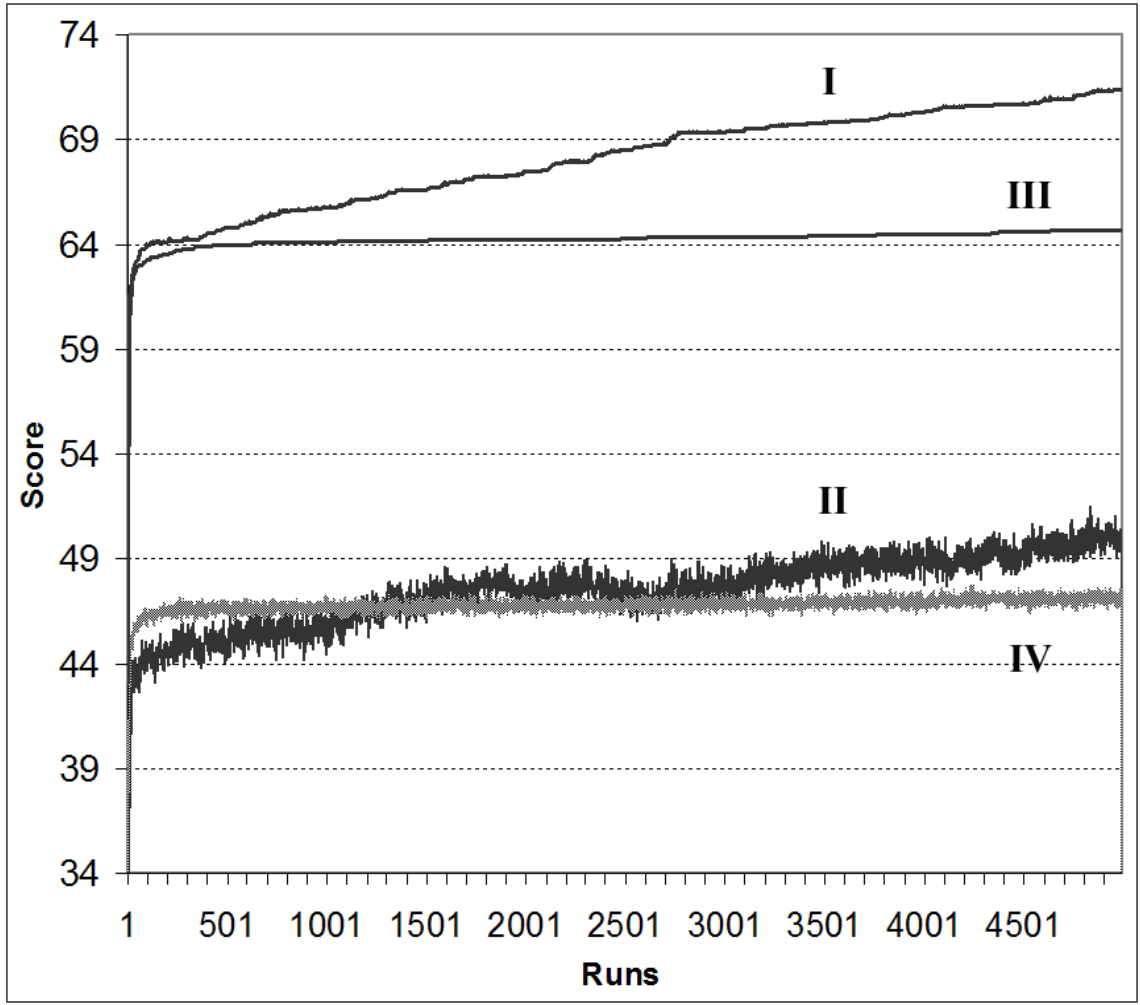

Figure 7. Comparison of the learning abilities of ant populations in regard to transposon length. The curves of maximum and mean score dynamics (the transposon sequence size parameters were 5 to 32), averaged over 100 runs are compared with the corresponding curves, presented on Figure 5, where the transposon sequence size parameters were 5 to 11. The values of all other parameters are the same as in Figure 5. **I** are the best-of-generation scores and **II** are the mean scores for the large transposon runs; **III** are the best-of-generation scores and **IV** are the mean scores for the runs from Figure 5.

The analysis of transposon population dynamics has shown that their observed general features are the same for both cases - transposons with a high upper size threshold as well as transposons with a low one. In several generations, transposons of maximum possible length begin to dominate. As may be inferred from the diagram (Fig. 7), transposons of length 32 are more effective mutators than transposons of length 11. It seems that the larger the transposon becomes, the more intelligent is it as a mutator. We cannot give any simple explanation of this effect so far.

## 3.4. Horizontal Transmission of Transposons Is Necessary For Their Effective Mutation Effect

Because transposons are transmitted vertically, that is, from ancestors to descendants, transposons belonging to a host that had a reproductive advantage would rapidly spread through a population and develop new forms in the process. But this process is insufficient per se for effective acceleration of ant learning. The two-place MGE operator, performing horizontal distribution of transposons from one ant to another is necessary for an increase of ant training ability. In Fig. 8 we illustrate the results of comparing the test presented in Fig. 5, with a similar test, in which the frequency of applying the two-place MGE operator was reduced by a factor of 10 to 5%. This parameter determines the percentage of the ant population that is subjected to the action of the two-place MGE operator in a generation. In all previous experiments presented in this article, this percentage was 50%.

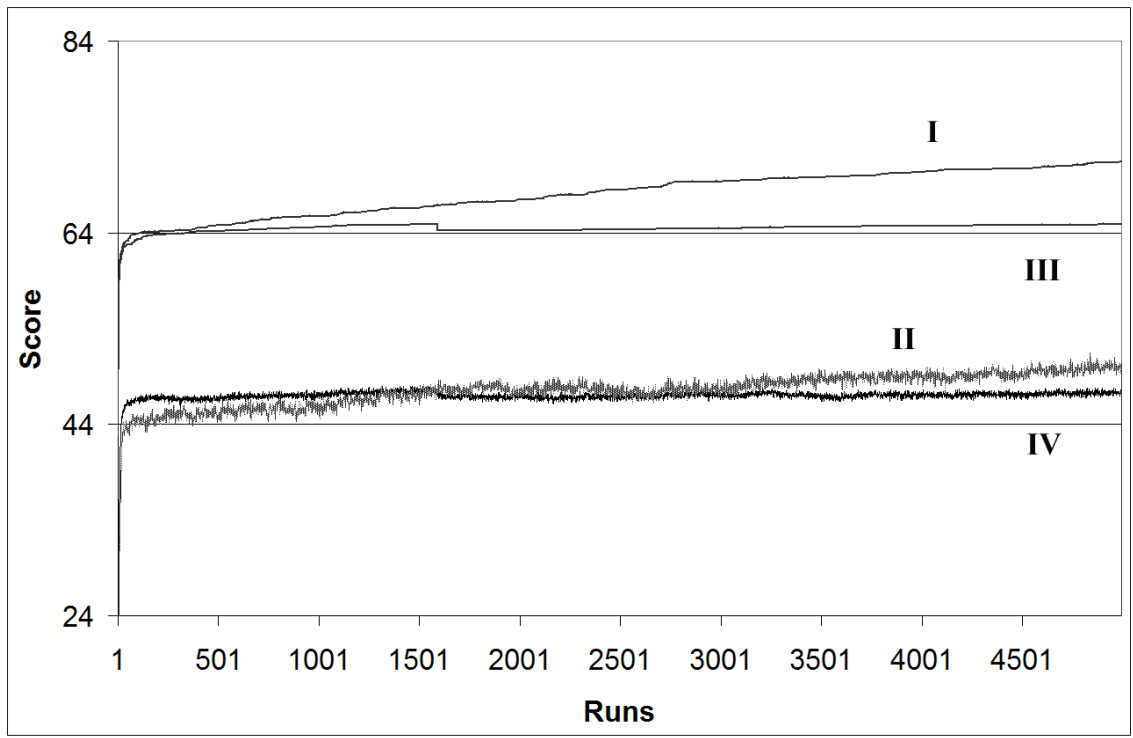

Figure 8. The influence of decreasing two-place MGE operator application frequency on ant learning abilities. The other parameters are the same as in the previous experiments (see caption to Figure 5). **I** are the best-of-generation scores and **II** are the mean scores for the test runs; **III** are the best-of-generation scores and **IV** are the mean scores for the test runs from the Figure 5.

That an obvious decrease in ant learning ability coincides with the decrease in the frequency of two-place operator application is evident from the diagram. It lowers the efficacy further and makes it almost equal to the control (the case without transposons).

## 3.5. Ant Populations Adapt to the Selective Pressure of Transposons

In contrast to the external, rigidly predetermined operator of mutation in classical GA, intelligent mutators co-evolve with their host. The maximum length of a transposon can be critical to the fitness of the host, as a transposon that takes up too much space in the host's genome and yet does not positively contribute to the navigation algorithm will be detrimental to the survival of the host.

Small sized ant populations are predisposed to noticeable fluctuations, what made them inappropriate for our analysis. For this reason, in this section we presented the results of computer experiments with a more abundant ant population of 1000 individuals. The large population is more stable, and as such, the fine peculiarities of the various dynamics are obvious (Fig. 9).

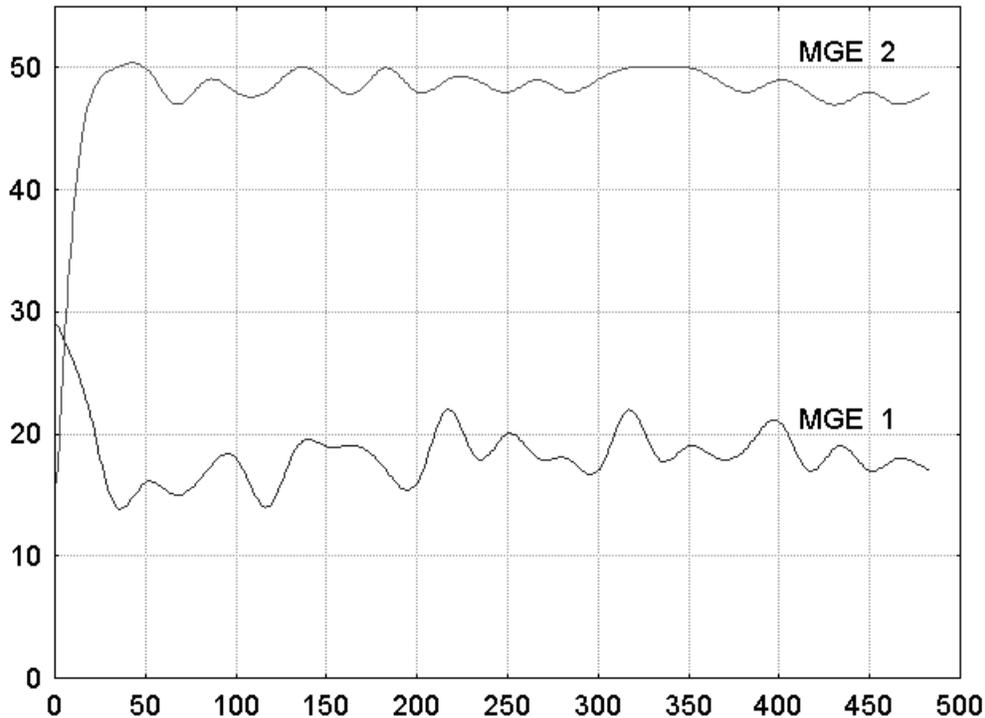

Figure 9. The dynamics of the number of individuals in the ant population (in percentages), subjected to the action of one-place and two-place MGE operators (MGE 1 and MGE 2, correspondingly).

The analysis of all computer experiments, carried out and presented in previous sections, made it possible to propose the following model of the transposon's role in ant population dynamics.

In accordance to our model, transposons subject the population of hosts (i.e., ants) to selective pressure for only a short period of evolutionary time (as a maximum, some tens of generations). So, as can be seen from diagram 9, the percentage of the ant population subjected to the action of the one-place and two-place MGE operators numbers 17% and 48% respectively, over 40 - 50 generations. For this short period of time, the ant population is able to adapt to the transposons' selection pressure at the expense of loosing short transposons with length of less than 11.

## *4. DISCUSSION*

The problem of programming an artificial ant to follow the Santa Fe trail has been repeatedly used as a benchmark problem in GP [10, 12, 16, 21, 22, 25, 28]. Recently Langdon and Poli have shown that the performance of several techniques is not much better than the best performance obtainable using uniform random search [23]. According to these authors, the search space is large and forms a Karst landscape containing many false peaks and many plateaus rived with deep valleys. The problem fitness landscape is difficult for hill climbers and the problem is also difficult for GA [23].

There are many techniques capable of finding solutions to the ant problem (GA, GP, simulated annealing, hill climbing) and although these have different performance characteristics, the best typically only do marginally better than the best performance that could be obtained with a random search [23]. For this reason, the ant problem may be indicative of real optimization problem spaces and as such, may be worthy of further study.

## 4.1. Dominant Transposon Forms Are the Components of Effective Navigation Algorithms

Let us discuss a possible evolutionary process that leads to the emergence of a more general algorithm for artificial ant pathfinding on trails of high difficulty and arbitrary organisation. The initial section of the trail in Figure 4 is organised as follows: a turn to the right, a turn to the left, and a gap two cells in length. A population of 1000 individuals developed an effective algorithm for traversing the first half of the trail within a period of several generations. The algorithm is as follows: if there is no trail ahead, the ant turns to the right. If the trail is still not found, the ant makes two more turns to the right. If after this it still does not find the trail, the ant takes one step forward and "sniffs around" the adjacent cells. If it still does not find the trail, the ant takes one more step forward in the direction of the interrupted path and repeats the previous procedure. This is the RLLRF (or LRRLF) loop, first described in Section 2.4.1.

The ant may, however, encounter situations of a higher order of difficulty. For instance, the next cell of the trail may be four to five cells away diagonally from the last, as is the case of the gap between cells #64 and #65 on our trail in Figure 4. This becomes a "bottleneck" for the evolutionary process. It is clear that a population can learn step-by-step on a trail of gradually increasing complexity. But an approach where the population of ants generates a more general pathfinding algorithm on the basis of particular rules developed earlier (namely, RLLRF or LRRLF loops), is of much greater interest for us. This "intellectual" work is a result of active internal processes in an ant's genotype that lead to navigation rules spontaneously being made redundant.

In the case of an encounter with a trail element of high difficulty that the population was not trained to handle in an earlier section of the trail, the following course of events may occur. As the ant population is unable to traverse the "bottleneck", no ant in the population will be able to amass a higher score than others, and as such, no ant will have any evolutionary advantage over another. Thus, there will be no selection pressure in the population to overcome the obstacle. However, the internal population of transposons have their own selection pressure - i.e. to increase in size. This selection pressure will result in the simple algorithms mentioned earlier (namely, RLLRF and LRRLF loops) being replaced by an increasing number of bulkier programs (see Table 1) through the mechanisms of transposon growth and transposon mutagenesis (See Sections 2.4.1.1 and 2.4.1.2).

These bulkier programs are able to solve the initial navigation problem. However, they do this with obvious redundancy. Specifically, new programs appear when the ant is sniffing around the surroundings not directly ahead in the direction of the lost trail, but moving zigzag in the same direction:

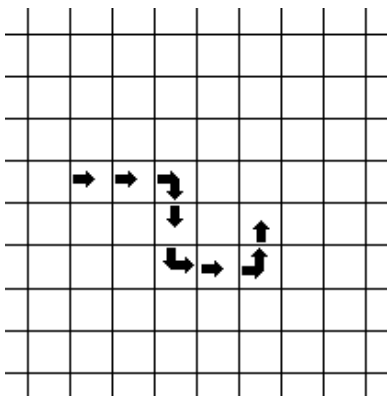

Redundant programs like this can be applied to more complex navigation. If the population is still in a state of stagnancy, than eventually navigation programs will appear with high redundancy due to recursion of the bulkier algorithm. This can happen because of the periodicity of transposon code. Some of these redundant programs, for example, will provide instructions for sniffing around the surroundings in the direction of the lost trail, but by a closed contour, which will catch a continuation of trail five cells to the right and five cells up:

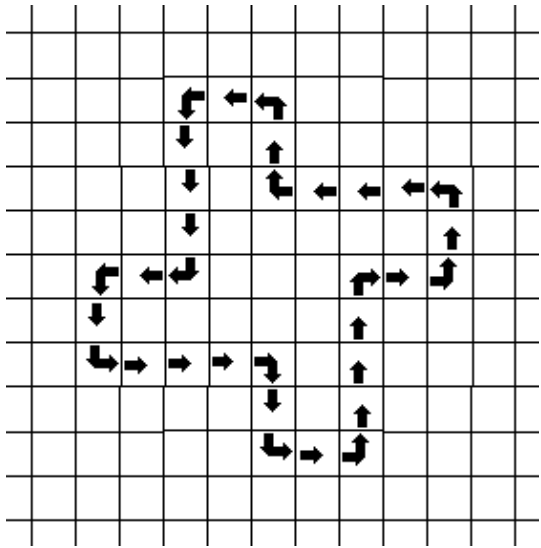

As a result, the evolutionary process will develop a more general algorithm of navigation, as is the case of the examples in Table 1. It is important that these algorithms provide instructions for searching through the surroundings in the vicinity of the last cell in the discovered trail. In this sense these algorithms are general. Ants of this type are able to pass through trails with chaotically scattered difficult elements.

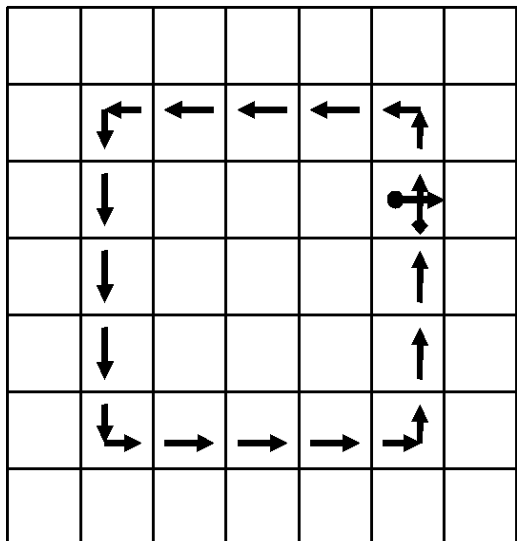

Figure 10. Example of the closed trajectory produced by four-fold execution of the transposon code 1 from the Table 1. 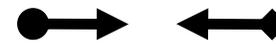 are the beginning and the end ant positions, correspondingly. The directions of the arrows are indicative of the orientation of the ant.

The results of careful analysis of the organization of several tens of dominant transposons (11-13 elements in length), taken from those ant populations that coped with the navigation task, can be summarized as follows:

1) By definition, the transposon-program begins and ends with the zero state, i.e., it is a loop, executed over and over until the ant meets a black cell. Typical examples of transposon-programs and corresponding ant movement trajectories are given in Table 1.

2) Four-fold execution of the transposon-program produces in most cases closed ant trajectories, i.e., the ant will return to the starting position. Examples of components of these closed trajectories are given in Table 1. As a rule, the closed contour is located in domains 4×4 or 5×5 cells in size (Fig. 10). Some cases of transposon-programs are encountered which provide ant movement back and forth along the polygonal path (See Table 1, code 2).

3) Ordinarily, the transposon program begins to work not from the zero state but from the $n^{th}$ state, which is specific to every transposon. This transition into the $n^{th}$ state takes place as soon as the ant (the host of the transposon) runs against a white cell.

4) Starting the transposon-program from the $n^{th}$ state allows the execution of the simplest navigation algorithm, necessary for overcoming the simplest gaps that are arranged in the first half of the trail ("looking around", then one step ahead, "looking around" again and so forth). This algorithm permits the successful passage of the trail up to the $64^{th}$ cell inclusive (Table 1, code 1 and 4, RLLRF and LRRLF, correspondingly).

5) The majority of transposon-programs guarantee the overcoming of the element of high complexity between the $64^{th}$ and $65^{th}$ cells.

6) Some transposons are not suitable for navigation programs. In that case, the chromosome elements located in the transposon-free domain take control of navigation.

Table 1. Examples of ant paths determined by transposons.

| No | Transposon's code | Ant's path |
|---|---|---|
| 1. | 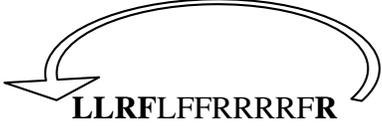 **LLRF**LFFRRRR**FR** | 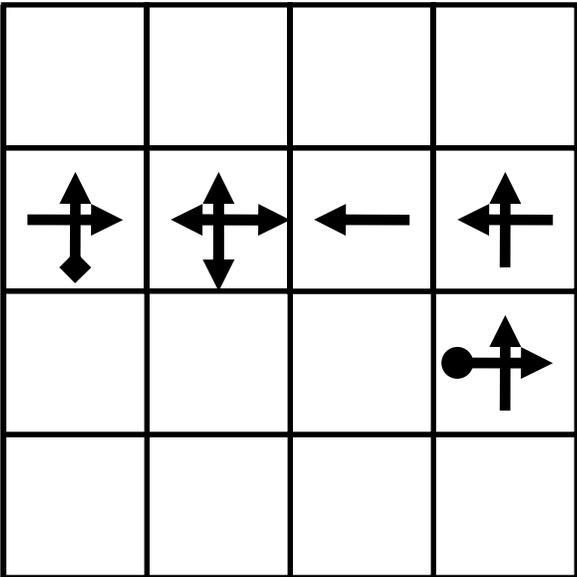 |

| 2. | ↺ LLRFLFFLRRFL | 4×4 grid with arrows |
|---|---|---|

| 3. | ↺ RRLFRFRFFLFF | 4×4 grid with arrows |
|---|---|---|

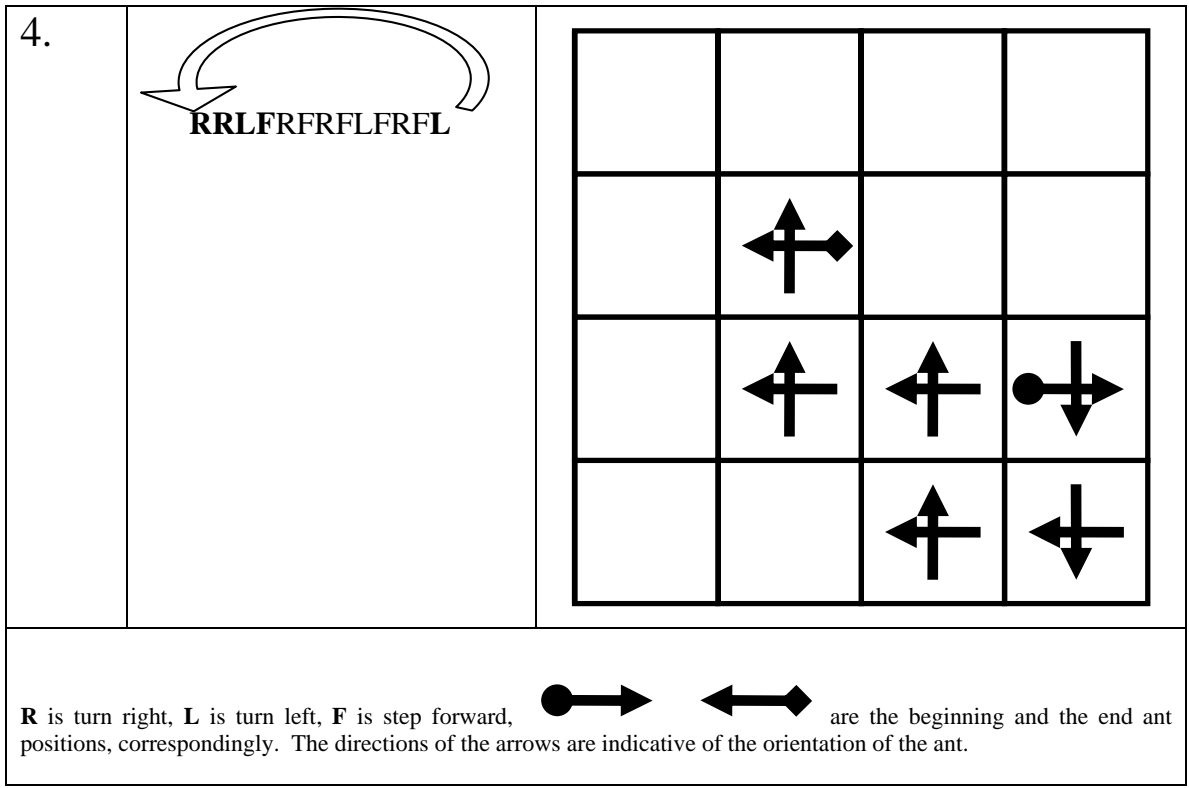

| 4. | **RRLF**RFRFLFRF**L** | |

R is turn right, L is turn left, F is step forward, ●→ ←◆ are the beginning and the end ant positions, correspondingly. The directions of the arrows are indicative of the orientation of the ant.

Detailed analysis of the organization of dominant transposon forms in ant populations that succeeded in finding an effective navigation program showed that the transposons themselves become components of these programs. The case in point is the portion of the navigation program that is used for effective "snuffing around" in a situation where the ant is faced with a wide gap.

In other words, we have determined that the transposons in our experiments take an active hand in the propagation of their hosts' building blocks.

## 4.2. Intelligent Mutators Have a Search Space Confining Effect

The Muir's Trail search space has a rugged geometry due to the specific and discrete characteristics of the problem. That is why gradient methods are not effective here [23]. Moreover, this ant navigation problem is classified as a GA hard problem, especially if the trail is not designed specifically for ant population training. The efficiency of transposons in the role of intelligent mutators can be measured by their search space domain confining ability. Therefore, the selection criteria inserted into MGE operators had to increase the probability of effective navigation algorithm discovery on the elements of high complexity. We proposed that quasi-periodical state successions that meet the definition of a transposon will be picked up by selection and used as a component of a universal navigation algorithm.

A comparison of mutation frequencies in the experiment and control with the corresponding learning rates confirm a multiple reduction of evaluation numbers needed for reaching the same required learning rate in experiments with transposons. The mutation frequencies for control runs (Fig. 5) have a crossover rate + mutation rate = 0.0001+0.04 P/bit/generation; MGE1 and MGE2 operators add on average 0.0027 and 0.0075 P/bit/generation respectively. In other words, transposons on average add 0.012 P/bit/generation to the control value of 0.041. This addition brings to multiple acceleration of ant population learning. Hence, according to fig. 6, up to the end of the experiment (4622nd time-step) the control set gives a max score of 6.47, whereas in the test set this value is already attained on the 451st time-step, i.e. 10 times faster.

And what is more, though the ant test is now accepted as a classical one, its implementation in the form of a FSA is far from optimum. Firstly, the 32 states are obviously too many for the classical trails. This redundancy brings an enormous extension of search space dimensionality. Secondly, the possibility of generating various action sequences in the situation when the ant is faced with a black cell is redundant as well. It is clear that the "step forward" is the only adequate action in this situation. Thirdly, the action NOP is useless.

Apparently, the simplest and most effective method of this type of navigation problem solving would be using symbol strings, composed of combinations of three possible actions (FWD, RGT, and LFT) in the situation when the ant faces a white cell. Applying the MGE-approach to the ant problem, we essentially perform this kind of search space size reduction. In so doing, we deal only with the "white half" of the search space which is formed by all sequences of actions, executed while the ant has white cells in front of it. Furthermore, we escape internal loops, and exclude the NOP action. We suppose that this mechanism explains the MGE-approach effect i.e., up to ten-fold increase of evolutionary search rate. The additional contributing factor for the MGE-technique's vast increase in efficiency over the control is the inclusion of the transposition procedure, the application frequency of which is substantially higher than that of the other mutation operators.

## *5. Future Work*

Besides the good results obtained with these MGE operators, we intend to do more empirical work to clarify some results in need of explanation. First of all, an exhaustive study will be make to find the exact relations between transposon population dynamics and ant population dynamics, namely whether "epidemics" of new transposon mutants are caused by a new effective navigation algorithm conveyed by this mutant (See the section 3.2), and why long transposons are more effective in facilitating ants' learning than shorter ones (See the section 3.2).

More generally, our future work will be aimed at improvement and generalization both animat as is and the MGE-algorithms. To do this we intend to extend our approach to other benchmark tests based on FSA-animats and, more generally, to other FSA-implemented problems.

In the case of the ant-pathfinder problem, we are going to use trails that truly require the usage of the ant's state memory in full scale. As far as we know, nobody yet paid attention to the fact that the ant problems for the classic "Santa Fe" and "Los Altos" trails don't need a large state memory. The only thing that a well-trained ant must remember is whether or not it is on the path. That is why these test tasks could be solved by many other approaches including those which don't use memory [23]. To force an ant to use its memory at full capacity, we need to test it against more sophisticated paths.

One of the natural ways to design such paths is to use some elements of the path as a signal to apply one of algorithms to follow next, large enough part of the path, until this part ends and another signal block from black cells will be faced. For example, if an ant is faced with this block of black cells in a sequence, then the next part of the path would be a scatter of single black cells, the most effective way to find which in sequence would be, for example, to cyclically use the algorithm FFRFL. In such a case, an ant will have to remember at a minimum dozen previous states / actions, in order to recognize and distinguish different blocks of black cells.

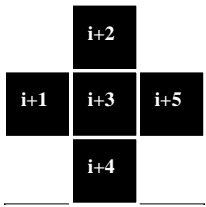

Classical animat for the ant-tracker problem have only one, frontal sensor [17]. Using more than one frontal sensor represents a substantial modification of the ant-tracker problem, but one that moves it closer to the goal of simulating an ant [3]. That is why we plan to use the FSA approach to design the multisensor animats, proposed by Ashlock and Joenks [4] and Ashlock and Freeman [3]. The agent senses its environment, consisting of the nearest eight cells, through 8 sensors. Ashlock and Freeman [3] did use an FSA implementation of such a multisensor animat and achieved surprisingly good results for the Tatrarus problem.

Two problems for virtual robots, the Tartarus and the Bulldozer [3 - 5, 57, 58], are similar to the ant-pathfinder problem and are good benchmarks for evolving agents that need state memory. We will extend our work on these problems in the FSA-implementation. The Ashlock and Freeman [3] FSA implementation of that 8-sensor animat makes it possible to apply with some modifications the artificial transposons and MGE-algorithms introduced in this paper. Though these two robotic problems are qualitatively more complicated then the ant-animat problem, we revealed that our transposon has an analogue in the Tartarus. Namely, Ashlock and Joenks [4] have proposed a way to control an agent's behavior by a variable length sequence of actions, which is repeated over and over until it runs out of moves. In this manner, an agent executes a dance, similar with the snuffling strategy of our ant.

We will also extend our work to the generalization of our MGE-algorithms. In this first attempt to apply new biologically inspired ideas, we exploited only one of many possible ways to implement it in the GA-ant pathfinder. For instance, this version of the MGE-algorithms exploit the decision table under the condition of zero-input only, after which it uses the rule to mutate the rightmost element of the transposon only, and the mutation mechanism is not completely random but generates, element-by-element, a periodic transposon sequence with the step being equal to 5. A natural generalization would be to give the evolutionary search the freedom to choose amongst all such possibilities.

To achieve this we are considering using a double-string implementation of a chromosome with an additional symbolic string to mark both transposon sequences (See Section 1.3) and to describe the details of the MGE-mutation rules (random mutagenesis or mutagenesis constrained by the periodicity with step equal N, mutagenesis targeted at the last transposon element or at any its elements, etc.) These generalizations might be especially effective in the case of highly complicated or skillfully devised tough versions of paths, just like we mentioned in the beginning of this section. The really hard versions of paths are especially interesting as benchmark problems close to real-life optimization problems, such as the evolutionary elaboration of tools for image processing, for instance.

That is why one more challenge for us in the problem of the evolution of virtual robots is the exploration of images. The ant-like animats have been used successfully for the task of image segmentation by evolving GP-ants [6]. Such extension of animat methodology makes it possible to automatically find low-level image processing tools, without making any assumption about the exploration strategy used. An animat fitted with simple sensors, able to move inside the image and to see the values of its surrounding pixels, is optimized by evolutionary search for performance of contour tracing, for example. The input is a natural grey level image: the animat's task is to find as many contour pixels as possible, using a minimal path length. The output is a binary image showing the pixels that the animat has detected.

However, this implementation of an animat has at minimum two serious faults, which are typical for any low-level image processing by animats [6]. Namely, the animat implements the most primitive, uneconomical algorithms, scanning, virtually pixel by pixel, the whole image. Besides that, the animat is not able to find all contour pixels and leaves gaps in the contour.

The high efficiency of MGE algorithms for the ant navigation problem is based on the elaboration of effective methods of "sniffing out" in nearest vicinity of a trail gap. Hence, it would be natural to apply this approach to improve the efficacy of contouring images with GA-animat.

**Acknowledgments**
This work was supported by INTAS grant No 97-3095. AVS is supported by Joint NSF/NIGMS BioMath Program, 1-R01-GM072022 and the National Institutes of Health, 2R56GM072022-06.